\documentclass[11pt]{article}
\usepackage[margin=1in]{geometry}
\usepackage{graphicx}
\usepackage{amsmath}
\usepackage{booktabs}
\usepackage{caption}
\usepackage{subcaption}
\usepackage[hidelinks]{hyperref}

\title{Architectural Trade-offs in Small Language Models Under Compute Constraints}
\author{Shivraj Singh Bhatti - ssbhatti@umass.edu}
\date{}

\begin{document}
\maketitle

\begin{abstract}
We present a systematic empirical study of small language models under strict compute constraints, analyzing how architectural choices and training budget interact to determine performance. Starting from a linear next-token predictor, we progressively introduce nonlinearities, self-attention, and multi-layer transformer architectures, evaluating each on character-level modeling of Tiny Shakespeare and word-level modeling of Penn Treebank (PTB) and WikiText-2. We compare models using test negative log-likelihood (NLL), parameter count, and approximate training FLOPs to characterize accuracy--efficiency trade-offs. Our results show that attention-based models dominate MLPs in per-FLOP efficiency even at small scale, while increasing depth or context without sufficient optimization can degrade performance. We further examine rotary positional embeddings (RoPE), finding that architectural techniques successful in large language models do not necessarily transfer to small-model regimes. Code and experiment logs: \href{https://github.com/shivraj-S-bhatti/language_model_architecture_comparison}.
\end{abstract}

\section{Introduction}
This work studies how architectural choices and compute budget affect the quality of small language models. We begin with a simple linear next-token predictor and progressively introduce nonlinearities, self-attention, and multi-layer transformer blocks \cite{vaswani2017attention}. All models are trained on Tiny Shakespeare (character-level) and evaluated using negative log-likelihood (NLL) on held-out test data. We then select the best-performing architecture and train word-level models on PTB \cite{marcus1993ptb} and WikiText-2 \cite{merity2016wikitext}. Throughout, we compare models in terms of test NLL, parameter count, and an approximate measure of training FLOPs to understand compute--performance trade-offs in a constrained regime.

\section{Experimental Setup}

\subsection{Datasets and Tokenization}
For the Tiny Shakespeare experiments we use the provided train/validation/test split (about 1\,MB of text in total). Tokenization is character-level: we build a vocabulary of all distinct characters across the three splits and map each character to an integer ID. PTB and WikiText-2 are treated as word-level datasets with whitespace tokenization and a fixed vocabulary built from the training split only.

Given a context length $T$ (``block size''), training examples for Tiny Shakespeare are constructed by sliding a window of $T$ characters over the corpus. Each example consists of an input sequence $x_{1:T}$ and target character $x_{T+1}$; models predict the distribution of $x_{T+1}$ given $x_{1:T}$.

\subsection{Model Architectures}
All Tiny Shakespeare models share the same input representation: characters are embedded into a $d_{\text{model}} = 128$ dimensional space via a learnable embedding matrix.

\paragraph{Linear model.}
The linear baseline flattens the embedded context into a vector of dimension $T d_{\text{model}}$ and applies a single linear layer to predict logits over the vocabulary. This is multinomial logistic regression over concatenated embeddings.

\paragraph{Multi-layer perceptron (MLP).}
The MLP model flattens the embedded context and passes it through two hidden layers with ReLU activations and dropout, followed by a linear output layer. Hidden dimensions $\{128,256,512\}$ are explored; the main configuration uses hidden size 256.

\paragraph{Self-attention model.}
The self-attention model adds learned positional embeddings to token embeddings and feeds the sequence into a single transformer-style block: causal multi-head self-attention (upper-triangular mask), a position-wise feedforward network, and layer norms with residual connections \cite{vaswani2017attention}. The final time step is projected to vocabulary logits. We sweep heads $\{1,2,4\}$; the main configuration uses 4 heads.

\paragraph{Multi-layer transformer.}
The transformer stacks $L$ blocks with embedding size 128 and feed-forward width 256. We sweep $L\in\{2,3,4\}$; $L=3$ is used as the primary model.

\subsection{Training Procedure}
All models are trained with Adam (learning rate $3\times 10^{-4}$, default $\beta$) and batch size 64. For Tiny Shakespeare we cap training positions per epoch at 50k for speed and use 10k positions each for validation and test.

Epoch counts are intentionally small because Tiny Shakespeare converges quickly in this regime. We train linear and MLP models for 3 epochs, and the attention and transformer models for 4 epochs. Validation loss is monitored each epoch and the best checkpoint is retained for testing and generation.

To compare models under different budgets, training FLOPs are approximated as
\[
\text{FLOPs} \approx 2 \times (\#\text{parameters}) \times (\#\text{training tokens}),
\]
which roughly corresponds to one forward and one backward pass per parameter per token. While coarse, this estimate is sufficient for relative comparisons across model families.

\section{Character-Level Modeling: Tiny Shakespeare}

\subsection{Learning Curves}
\begin{figure}[h]
  \centering
  \includegraphics[width=0.85\linewidth]{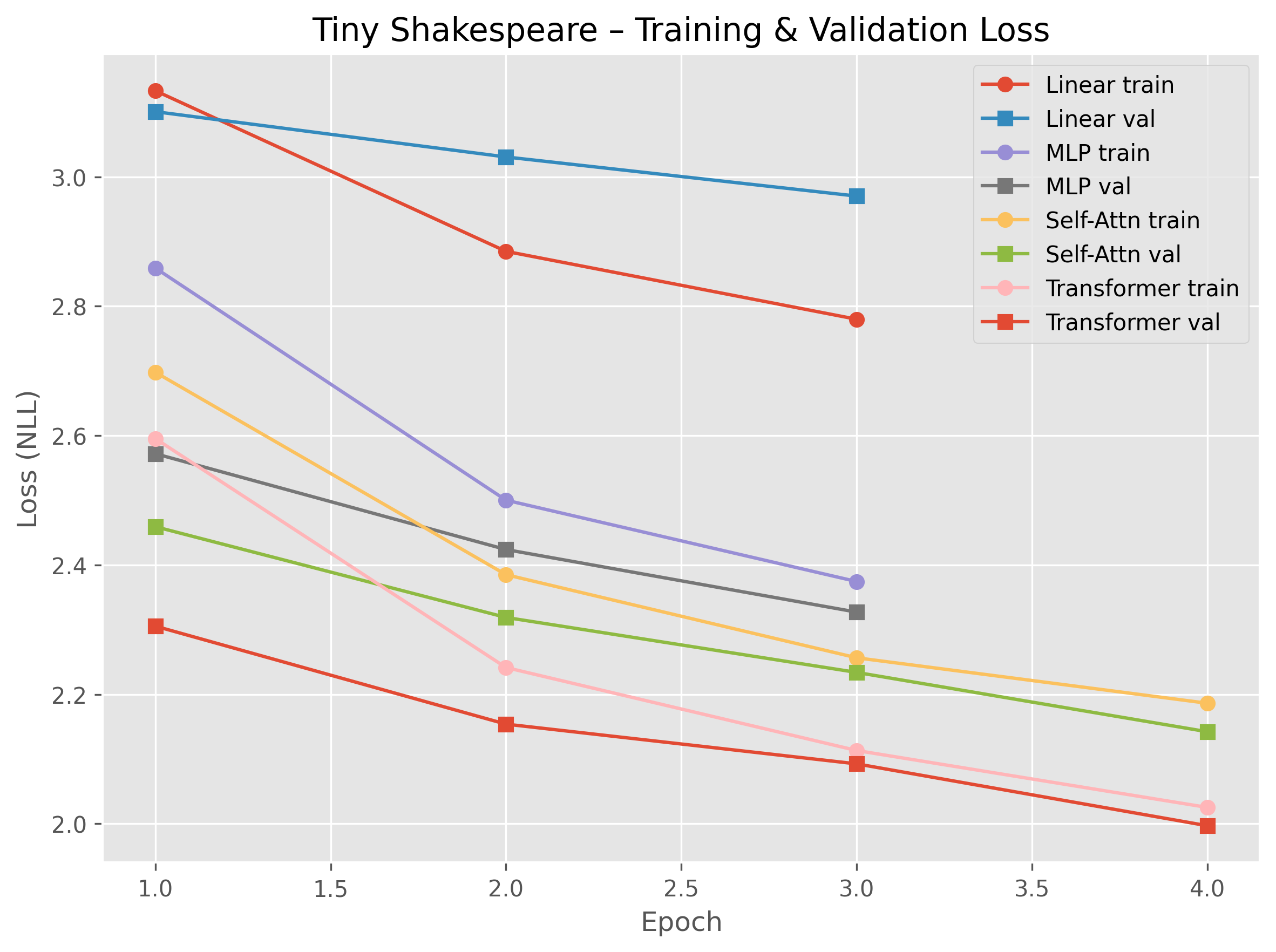}
  \caption{Training and validation NLL for Tiny Shakespeare across architectures. More expressive models achieve lower loss while still avoiding major overfitting in this setting.}
  \label{fig:tiny-train-val}
\end{figure}

Figure~\ref{fig:tiny-train-val} shows training and validation NLL across epochs for the four architectures.

The linear model reaches a test NLL of about 3.05. Adding nonlinear hidden layers in the MLP reduces test NLL to 2.32, and adding self-attention further improves it to 2.13. The multi-layer transformer achieves the best performance at roughly 2.01 NLL.

\subsection{Hyperparameter Sweeps}
We examine performance as a function of one design dimension per family, training each configuration for two epochs on a fixed subset for fast comparison.

\begin{figure}[h]
  \centering
  \includegraphics[width=0.6\linewidth]{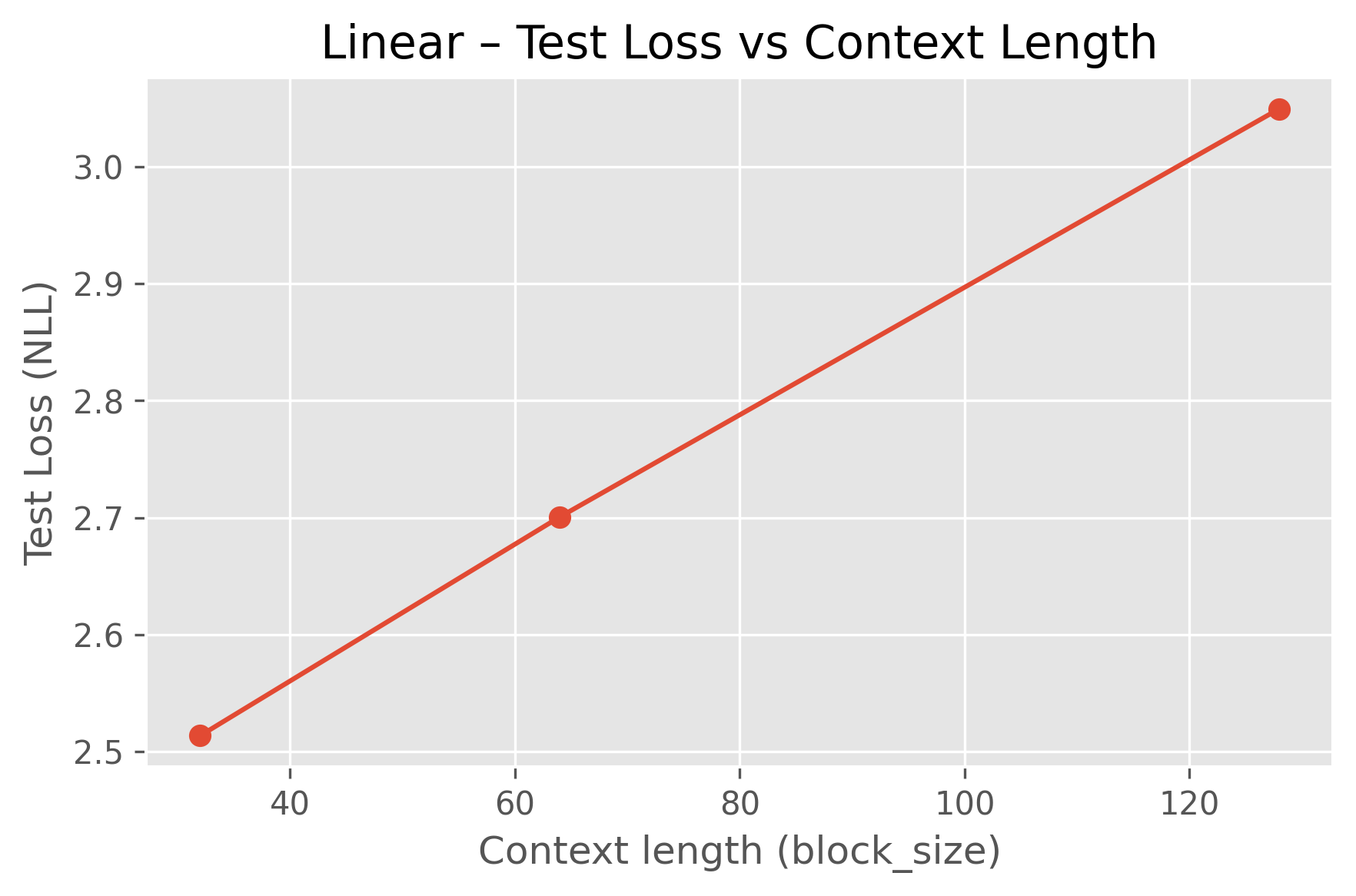}
  \caption{Linear model: test NLL vs.\ context length. Longer contexts increase dimensionality and parameters, but the model remains linear and appears under-trained at fixed optimization budget.}
  \label{fig:linear-sweep}
\end{figure}

\paragraph{Linear: context length.}
Figure~\ref{fig:linear-sweep} plots test NLL as a function of context length $T\in\{32,64,128\}$. The linear model performs best with the shortest context and degrades as $T$ grows, consistent with increasing parameter count without increasing modeling power.

\begin{figure}[h]
  \centering
  \includegraphics[width=0.6\linewidth]{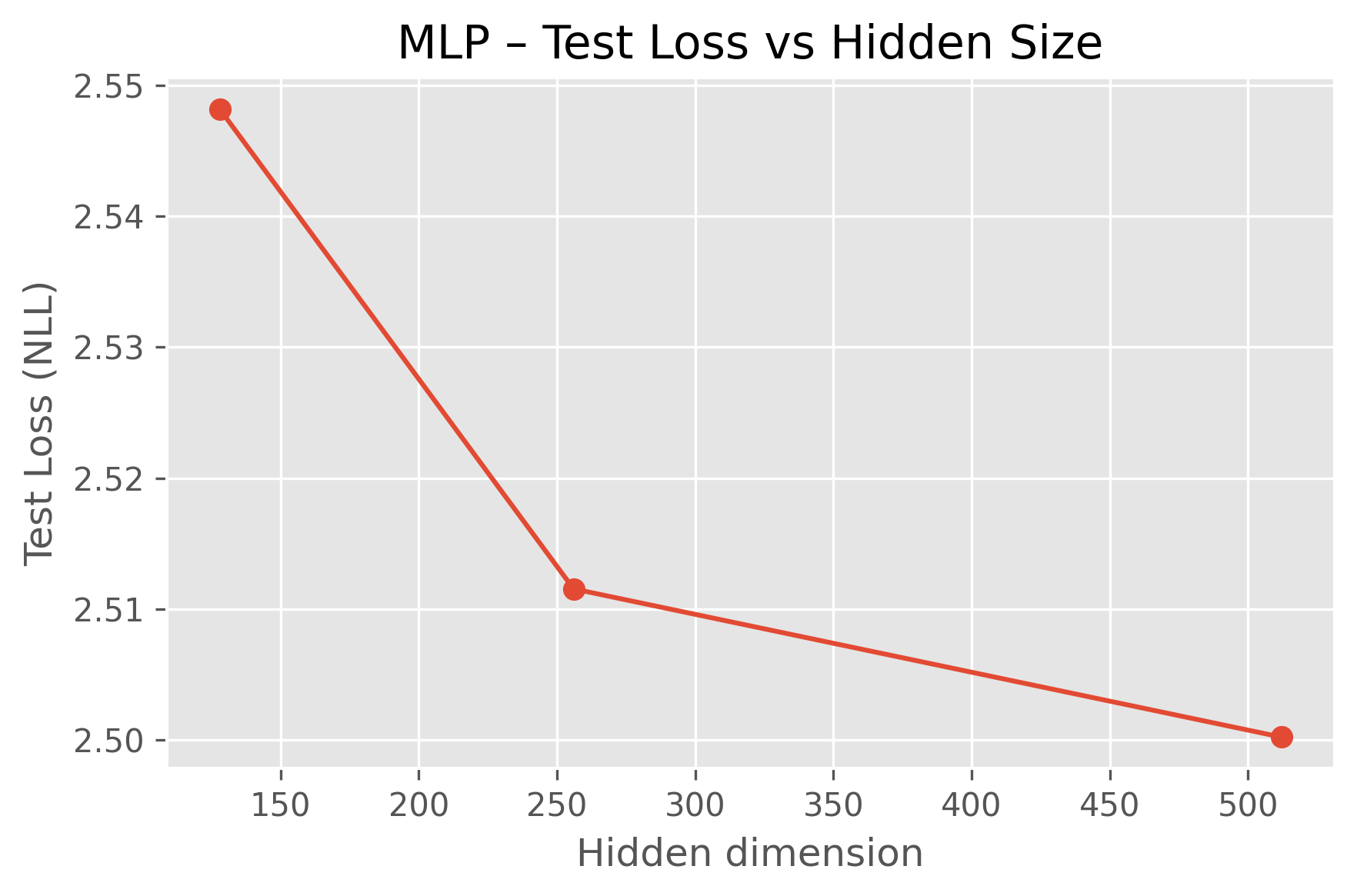}
  \caption{MLP: test NLL vs.\ hidden dimension. Wider layers help, but marginal gains shrink as parameter count grows.}
  \label{fig:mlp-sweep}
\end{figure}

\paragraph{MLP: hidden dimension.}
We vary hidden width $h\in\{128,256,512\}$ (Figure~\ref{fig:mlp-sweep}). Test NLL decreases with $h$, but gains diminish from 256 to 512, suggesting 256 is an efficient choice.

\begin{figure}[h]
  \centering
  \includegraphics[width=0.6\linewidth]{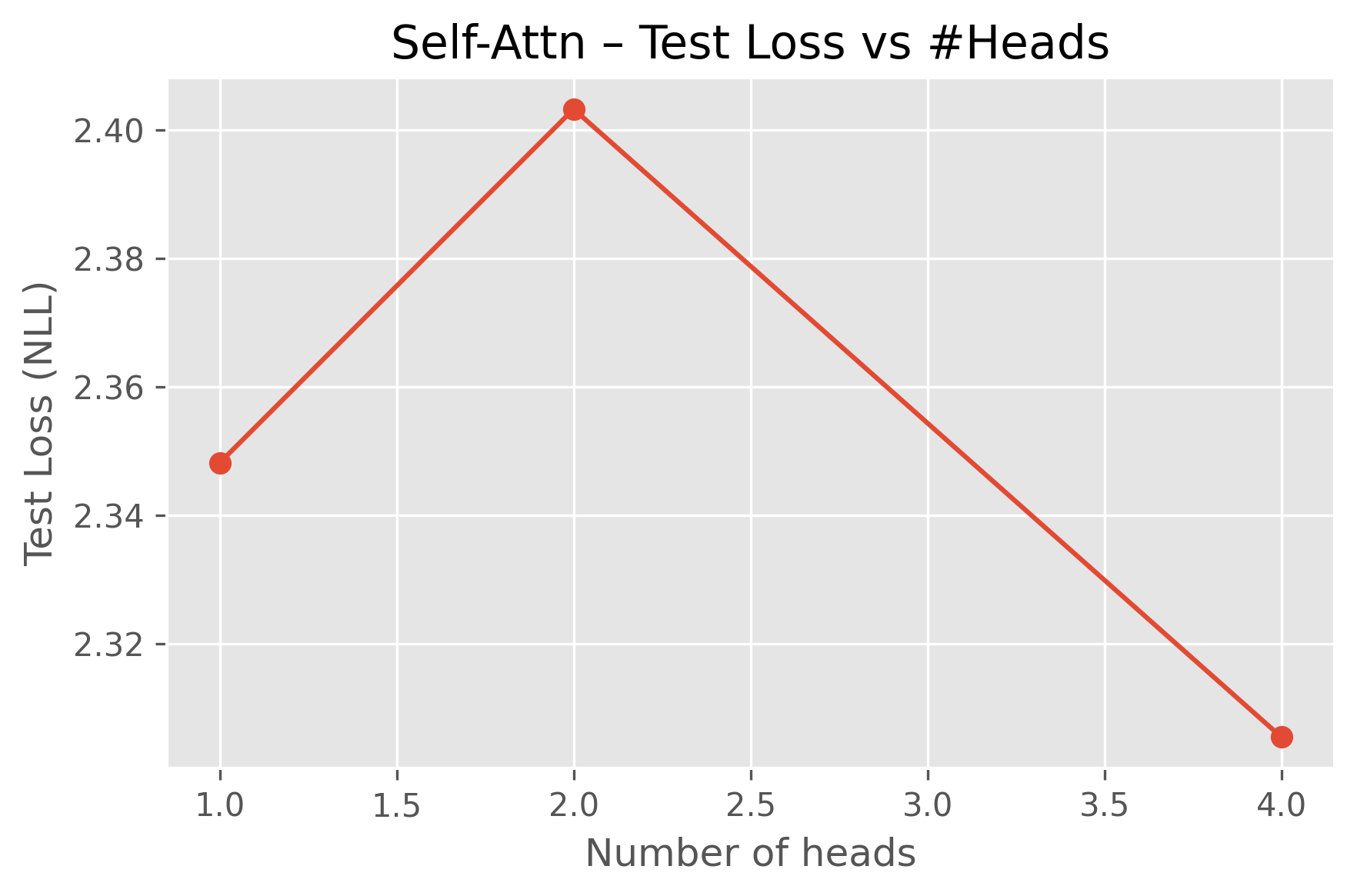}
  \caption{Self-attention: test NLL vs.\ number of heads. Four heads yields the best performance, consistent with learning multiple attention subspaces.}
  \label{fig:attn-sweep}
\end{figure}

\paragraph{Self-attention: number of heads.}
We vary heads $H\in\{1,2,4\}$ (Figure~\ref{fig:attn-sweep}). With fixed embedding size, 4 heads performs best in this regime.

\begin{figure}[h]
  \centering
  \includegraphics[width=0.6\linewidth]{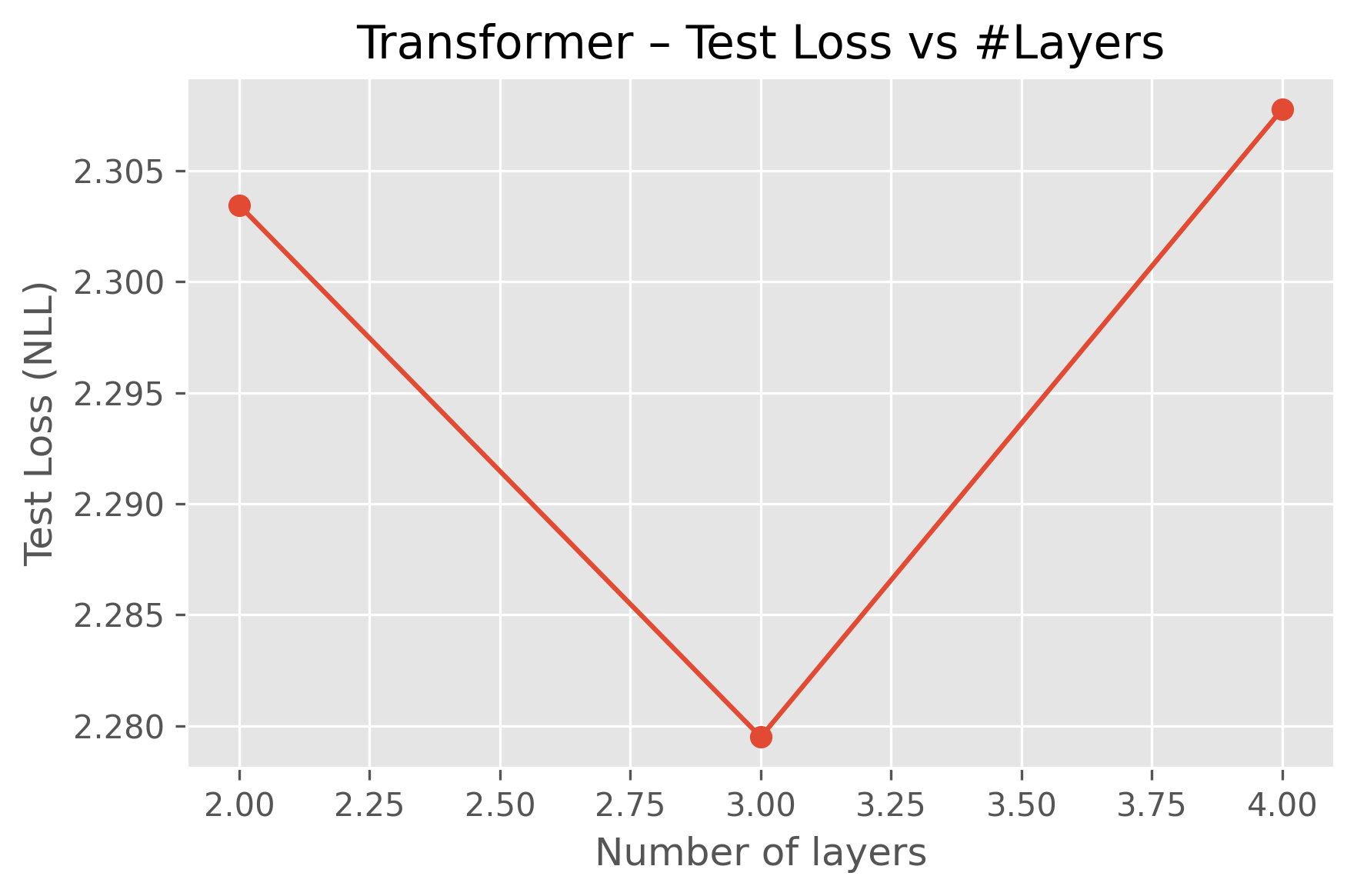}
  \caption{Transformer: test NLL vs.\ layers. Depth $L=3$ performs best here; $L=4$ likely suffers from insufficient optimization steps at fixed budget.}
  \label{fig:trans-sweep}
\end{figure}

\paragraph{Transformer: number of layers.}
We sweep depth $L\in\{2,3,4\}$ (Figure~\ref{fig:trans-sweep}). The 3-layer model slightly outperforms shallower and deeper variants, suggesting a good capacity--compute balance.

\subsection{Compute vs.\ Performance}
\begin{figure}[h]
  \centering
  \includegraphics[width=0.7\linewidth]{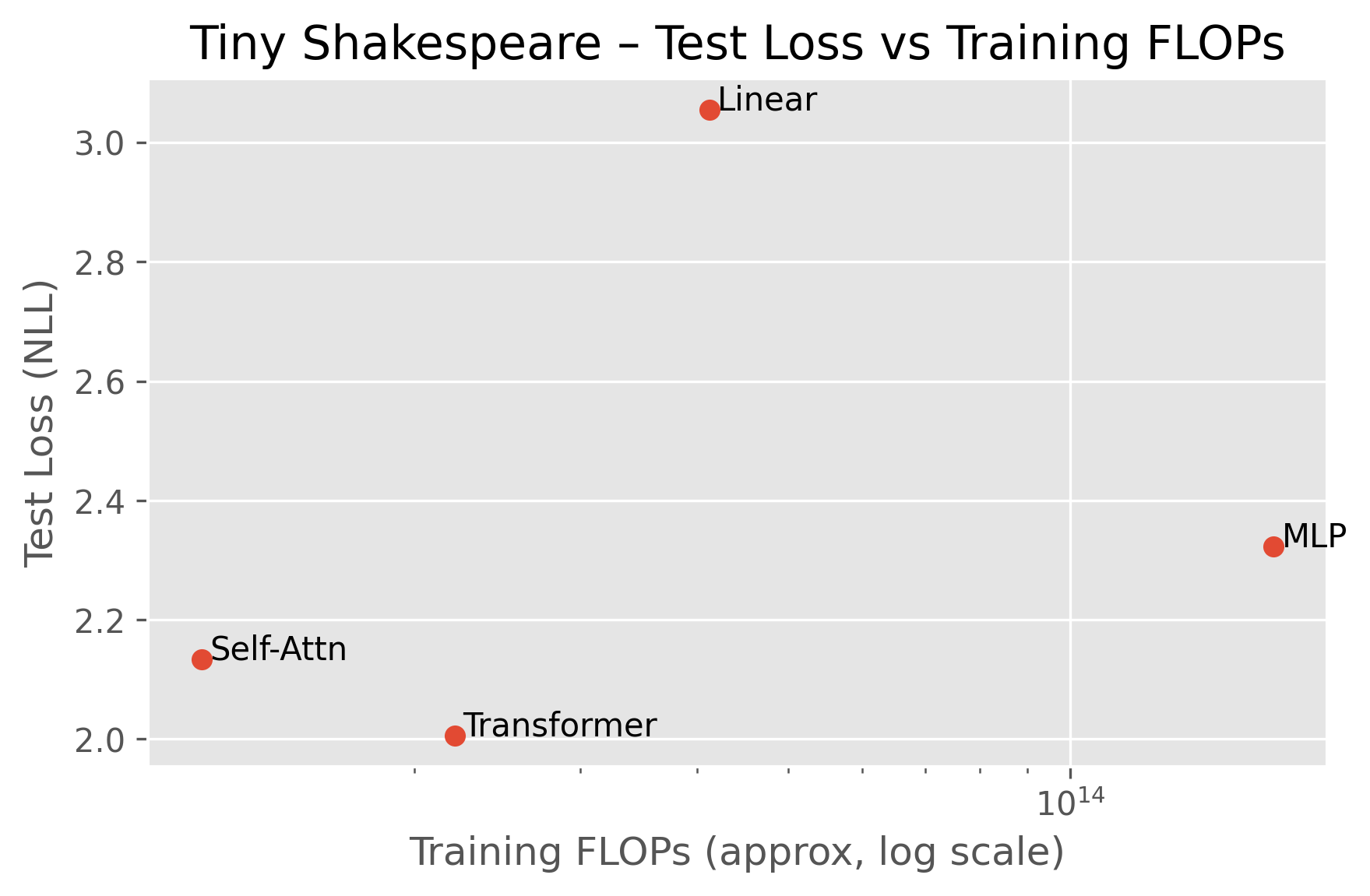}
  \caption{Test NLL vs.\ approximate training FLOPs (log scale). Attention-based models are markedly more efficient per FLOP than the MLP in this setup.}
  \label{fig:flops}
\end{figure}

Table~\ref{tab:baseline} summarizes Tiny Shakespeare configurations. Figure~\ref{fig:flops} visualizes test NLL against approximate training FLOPs (log scale).

\begin{table}[h]
  \centering
  \caption{Tiny Shakespeare models: parameter count, approximate training FLOPs, and test NLL.}
  \label{tab:baseline}
  \begin{tabular}{lrrr}
    \toprule
    Model & Params & Train FLOPs & Test NLL \\
    \midrule
    Linear       & 1{,}073{,}345 & $4.1\times 10^{13}$ & 3.05 \\
    MLP          & 4{,}285{,}377 & $1.6\times 10^{14}$ & 2.32 \\
    Self-Attn    &   231{,}617   & $1.2\times 10^{13}$ & 2.13 \\
    Transformer  &   430{,}785   & $2.2\times 10^{13}$ & 2.01 \\
    \bottomrule
  \end{tabular}
\end{table}

Two patterns stand out: (i) nonlinear models dramatically outperform the linear baseline, and (ii) self-attention/transformers provide better NLL per FLOP than the MLP.

\subsection{Qualitative Samples}
Each trained model generates a 100-character continuation of the prompt \texttt{HAMLET:}. Sample quality correlates with test NLL: the linear model produces scrambled characters, the MLP begins to capture word-like structure, and attention-based models produce clearer formatting (speaker tags, line breaks), with the transformer being most coherent.

\clearpage
\section{Word-Level Modeling on PTB and WikiText-2}
We reuse the 3-layer Tiny Shakespeare transformer (128-dimensional embeddings, 4 heads, feed-forward width 256) for word-level modeling on PTB and WikiText-2.

\subsection{Datasets and Tokenization}
We switch to word-level tokenization. Each corpus is read as whitespace-separated tokens; the vocabulary is built from the training split only. Vocabulary sizes are:
\begin{itemize}
  \item PTB: 9{,}999 tokens.
  \item WikiText-2: 33{,}277 tokens.
\end{itemize}
We use fixed context length $T=64$ words and construct examples by sliding a window over the token sequence. For efficiency, we cap training positions per epoch to approximately 80k for each dataset, and 20k for validation and test.

\subsection{Model and Training}
The architecture is unchanged except for the vocabulary-sized output layer. We train for 8 epochs using Adam ($3\times 10^{-4}$), batch size 32, dropout 0.1, and early stopping based on validation NLL.

\subsection{Training Dynamics}
\begin{figure}[h]
  \centering
  \begin{subfigure}{0.8\linewidth}
    \centering
    \includegraphics[width=\linewidth]{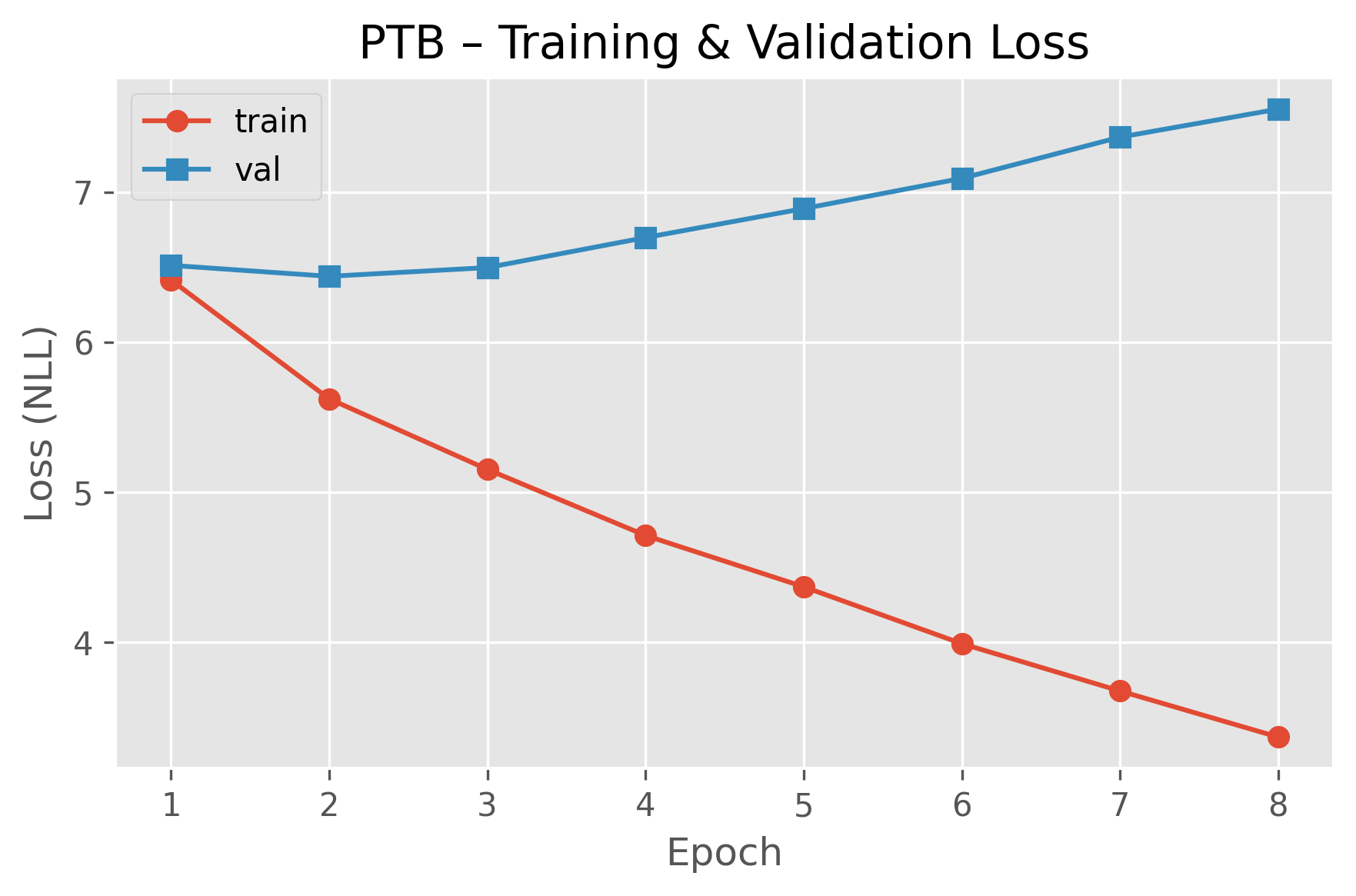}
    \caption{PTB}
  \end{subfigure}

  \vspace{0.3cm}

  \begin{subfigure}{0.8\linewidth}
    \centering
    \includegraphics[width=\linewidth]{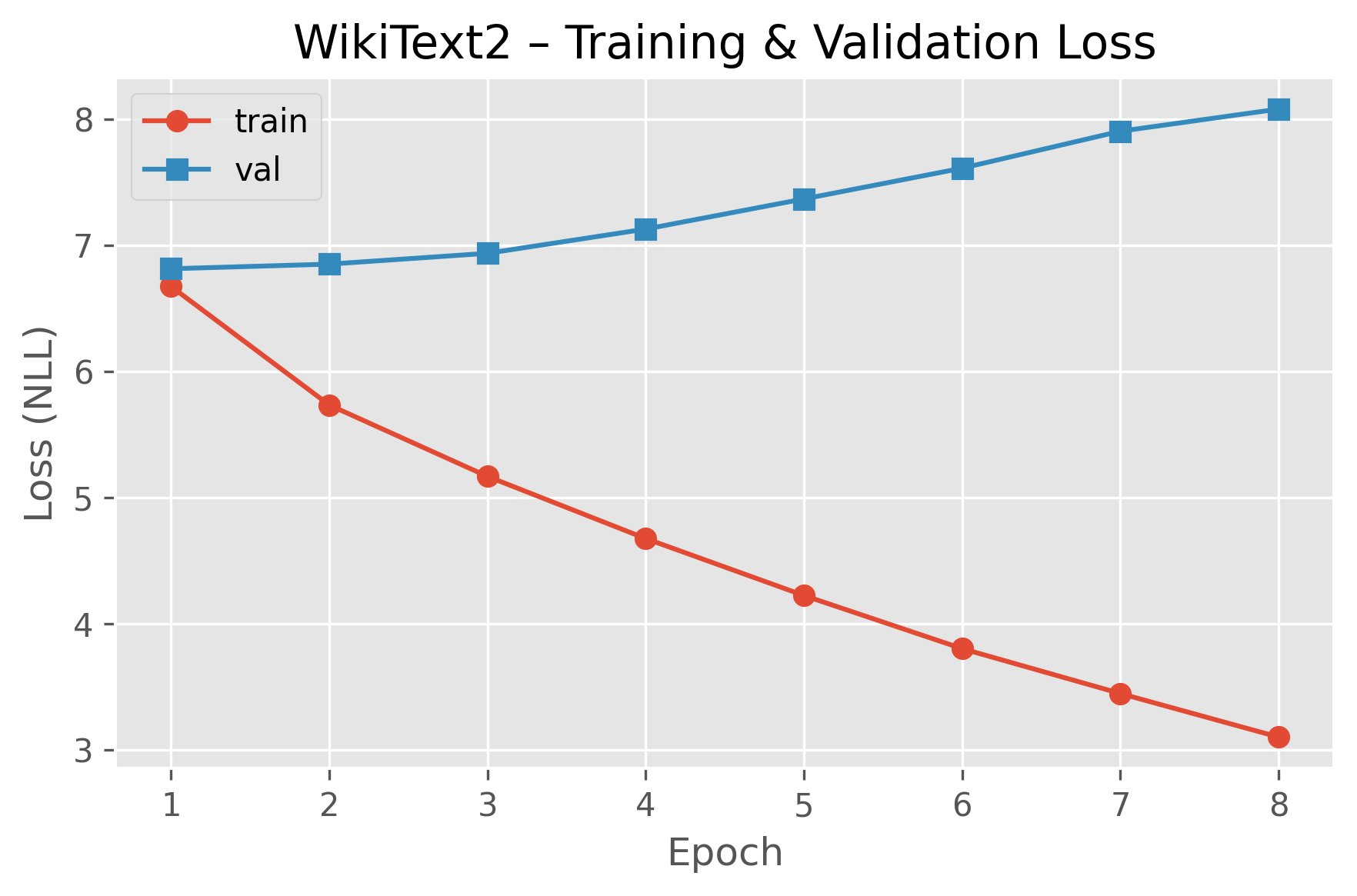}
    \caption{WikiText-2}
  \end{subfigure}
  \caption{Word-level transformer: training and validation NLL. Validation increases after early epochs, so the best checkpoint (minimum validation NLL) is used for test evaluation.}
  \label{fig:word-loss}
\end{figure}

Figure~\ref{fig:word-loss} shows training and validation NLL. Training loss decreases steadily, while validation loss bottoms out around epoch 2 and then increases, indicating rapid overfitting under limited regularization and short context.

Early-stopped test NLL is approximately 6.19 (PTB) and 6.89 (WikiText-2). The higher WikiText-2 NLL is expected because the corpus is more diverse and has a much larger vocabulary \cite{merity2016wikitext}.

\subsection{Compute vs.\ Performance}
We approximate training compute as
\[
  \text{FLOPs} \approx 2 \times \#\text{parameters} \times \#\text{training tokens} \times \text{epochs}.
\]

\begin{table}[h]
  \centering
  \caption{Word-level transformer: parameter count, approximate training FLOPs, and test NLL.}
  \label{tab:word-baseline}
  \begin{tabular}{lrrr}
    \toprule
    Dataset & Params & Train FLOPs & Test NLL \\
    \midrule
    PTB        & 2{,}975{,}631 & $2.44\times 10^{14}$ & 6.19 \\
    WikiText-2 & 8{,}958{,}077 & $7.33\times 10^{14}$ & 6.89 \\
    \bottomrule
  \end{tabular}
\end{table}

\begin{figure}[h]
  \centering
  \includegraphics[width=0.65\linewidth]{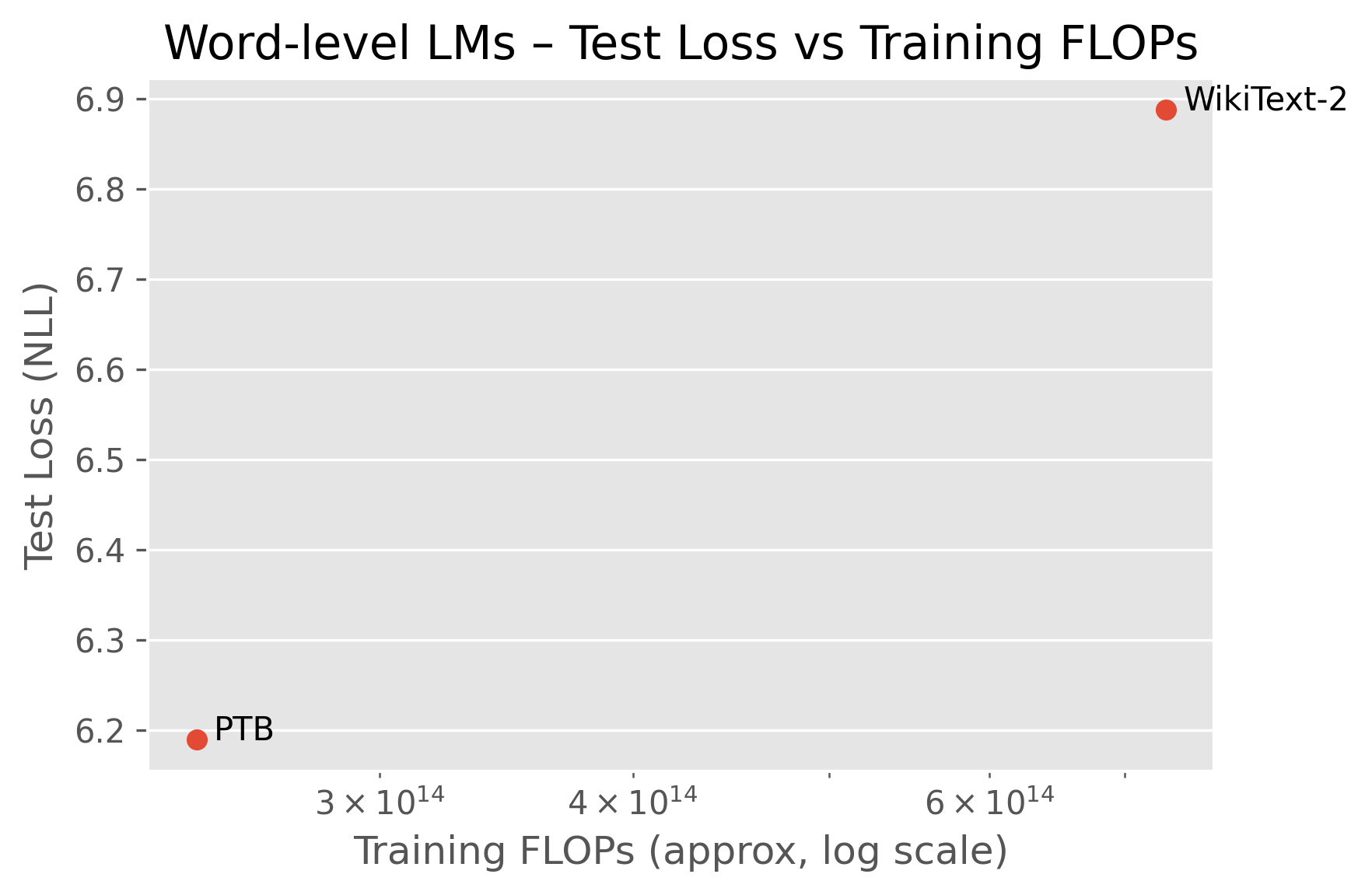}
  \caption{Word-level models: test NLL vs.\ approximate training FLOPs (log scale).}
  \label{fig:word-flops}
\end{figure}

\subsection{Qualitative Generations}
Qualitative samples reflect corpus style: PTB generations resemble financial news phrasing, while WikiText-2 generations resemble encyclopedia-like prose but lose semantic coherence over longer spans, consistent with the small model and limited context.

\clearpage
\section{Additional Experiment: Positional Encoding Transfer (RoPE)}
We evaluate rotary positional embeddings (RoPE) \cite{su2021rope}, a modern technique used in many contemporary transformers, by swapping only the positional encoding while keeping architecture and compute comparable.

We train two 3-layer Tiny Shakespeare transformers under identical settings:
\begin{itemize}
  \item \textbf{Learned positional embeddings} (baseline).
  \item \textbf{RoPE positional embeddings} (rotations applied to attention queries/keys).
\end{itemize}

\begin{table}[h]
\centering
\caption{Learned positions vs.\ RoPE on Tiny Shakespeare.}
\label{tab:rope-results}
\begin{tabular}{lccc}
\toprule
Model & Params & Train FLOPs & Test NLL \\
\midrule
Learned positional embeddings & 430{,}785 & $2.21 \times 10^{13}$ & 1.9738 \\
RoPE positional embeddings    & 414{,}401 & $2.12 \times 10^{13}$ & 2.0096 \\
\bottomrule
\end{tabular}
\end{table}

\begin{figure}[h]
\centering
\includegraphics[width=0.75\linewidth]{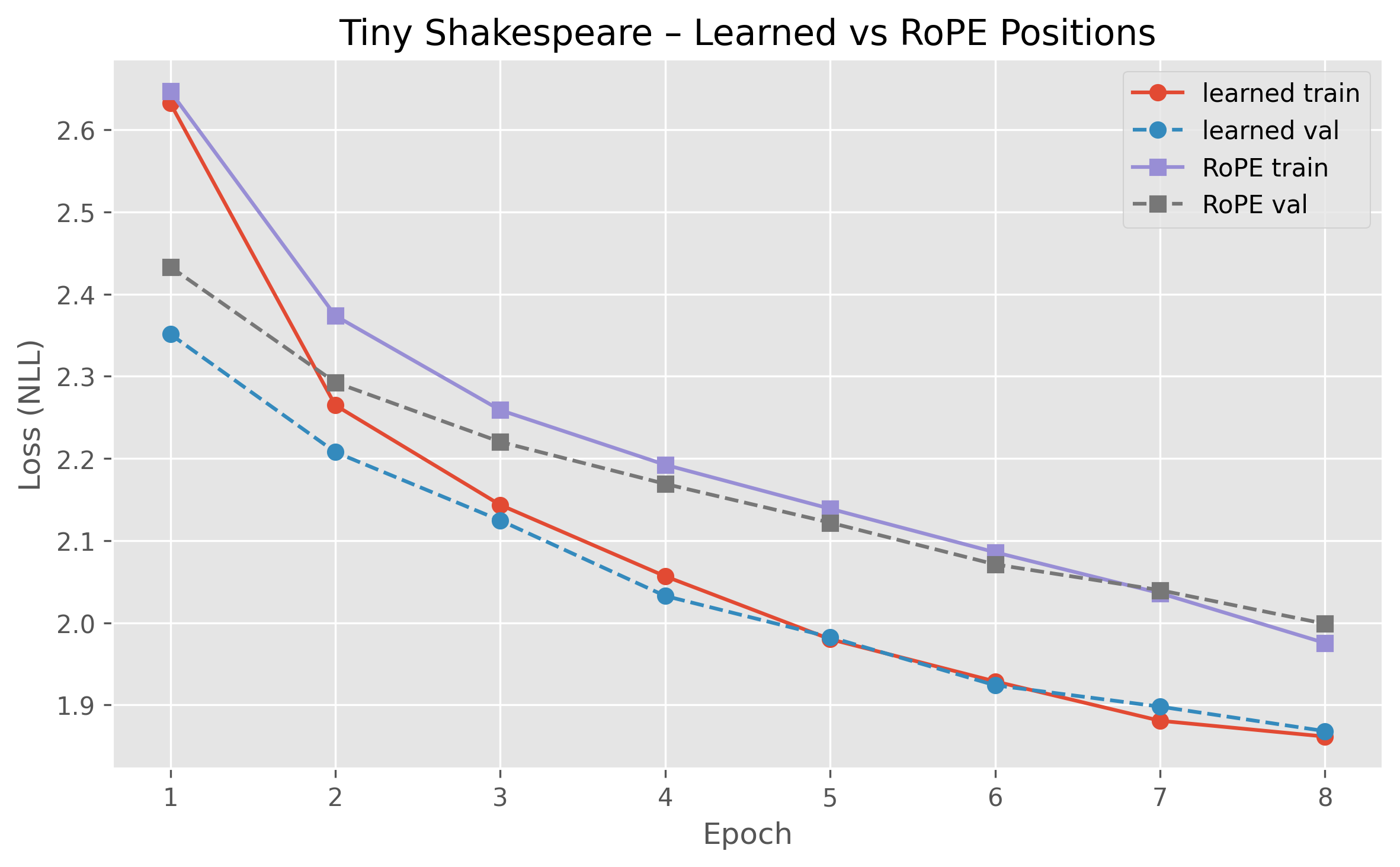}
\caption{Training and validation NLL: learned vs.\ RoPE positional embeddings.}
\label{fig:rope-loss}
\end{figure}

\begin{figure}[h]
\centering
\includegraphics[width=0.60\linewidth]{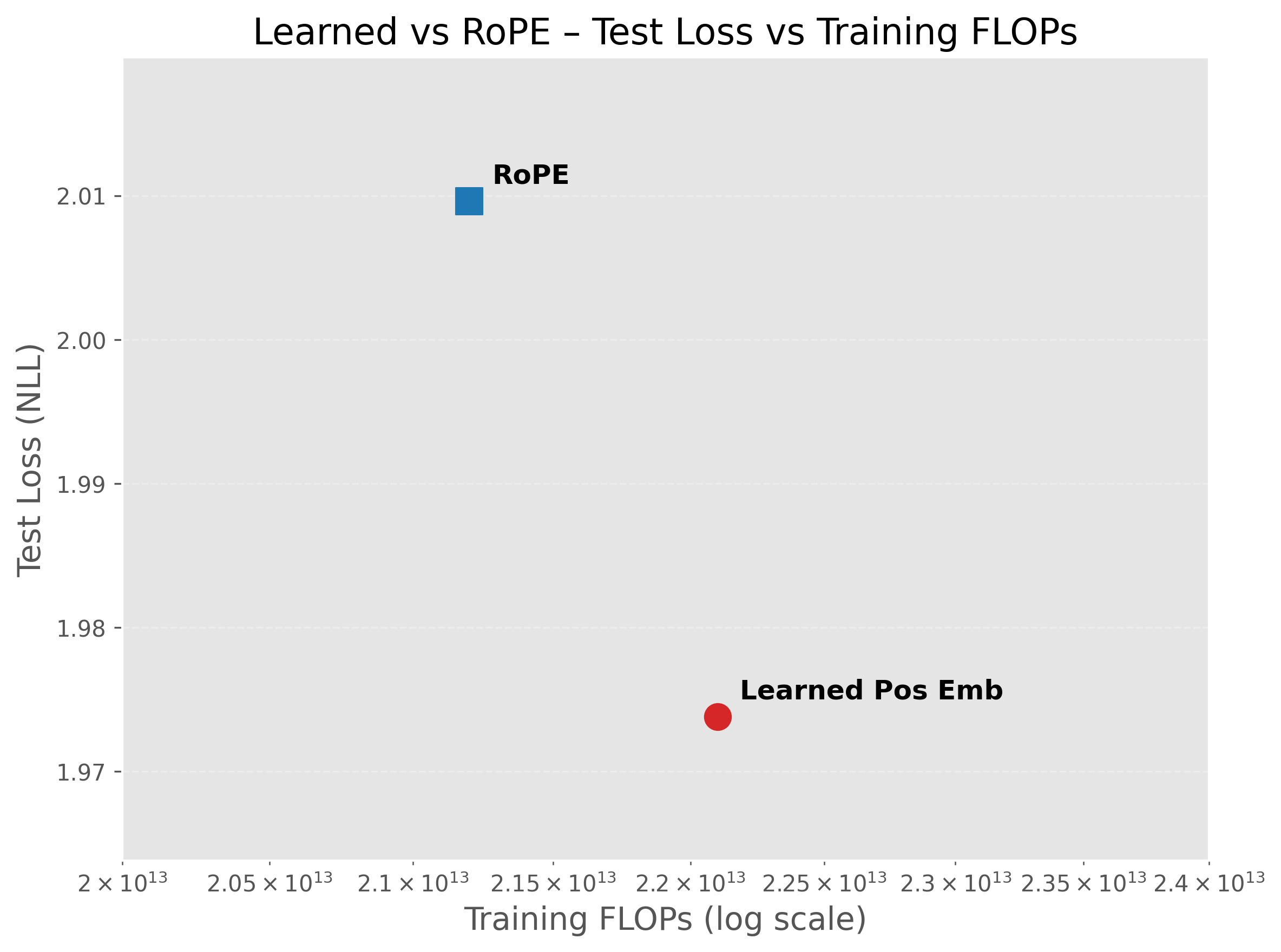}
\caption{Learned vs.\ RoPE: test NLL vs.\ approximate training FLOPs.}
\label{fig:rope-flops}
\end{figure}

RoPE slightly underperforms learned embeddings in this small-model regime. This suggests that techniques effective at large scale may not transfer downward without sufficient context length, data, and optimization budget. This negative result supports a broader theme of the study: architectural choices must be validated in the intended compute regime rather than adopted by default from large-model practice.

\clearpage
\section{Discussion}
Across Tiny Shakespeare, PTB, WikiText-2, and the RoPE ablation, a consistent picture emerges about small language modeling under constrained compute.

\textbf{Attention-based architectures are essential.} The linear model underfits, the MLP improves substantially, and attention-based models yield the best NLL per FLOP, highlighting the value of context-dependent representations over fixed concatenated features.

\textbf{Capacity must match compute.} Increasing context (linear) or depth (transformer) can hurt when training budget is fixed: additional parameters receive insufficient optimization steps. Hyperparameter sweeps reflect this trade-off.

\textbf{Word-level modeling magnifies scaling limits.} Reusing the Tiny Shakespeare transformer transfers surprisingly well in style, but validation loss rises quickly due to larger vocabularies, greater diversity, and short context windows.

\textbf{Modern LLM techniques may not help at small scale.} RoPE, while effective in large LLMs, does not improve test NLL here, emphasizing that ``best practices'' from large-scale training are not universally optimal.

Overall, architecture, scale, and compute must be considered jointly: effective small-model design comes from matching capacity to the available optimization budget and dataset complexity.

\section{Limitations}
FLOPs are approximated rather than measured precisely, and training uses capped token budgets per epoch for speed. The models are small and trained in short regimes, so emergent behaviors at large scale are out of scope. Nevertheless, the controlled setup cleanly exposes capacity--compute mismatches and per-FLOP efficiency differences.

\section{Summary}
We empirically compared a progression of small language models and found that attention-based models offer the strongest accuracy--efficiency trade-offs under constrained compute. Hyperparameter sweeps show that increasing context or depth without increasing optimization can degrade performance. The best Tiny Shakespeare transformer transfers across corpora for word-level modeling, but PTB and WikiText-2 reveal rapid overfitting and limited long-range coherence. A RoPE ablation further shows that techniques successful in large LLMs may not improve performance at small scale, reinforcing the need for regime-appropriate empirical validation.

\clearpage
\appendix
\section*{Appendix: Model-Generated Text Samples}
This appendix collects all model generations for reference (verbatim).

\subsection*{A.1 Tiny Shakespeare --- Learned Positional Embeddings}
\begin{quote}
HAMLET: \\
T: \\
u? \\
OPsetetbore as gq, \\
Henou,-afosneanvifofapredys Gvou, Ve?ese hinoke's yefme, \\
Ause! \\
I \\
Ce, oue, \\
\end{quote}

\subsection*{A.2 Tiny Shakespeare --- RoPE Positional Embeddings}
\begin{quote}
HAMLET: \\
Or Your my dease hat hour nother \\
Ibm bam pome Cleving that tood: I meart, \\
And tau meas cnoth anfoul \\
\end{quote}

\subsection*{A.3 PTB (Word-Level Transformer)}
\begin{quote}
the school announced that mr. signal task slow of school winter park markey by a husband
financial livestock sent a heavy departure machine nih international fernando growth
deal not copper the single talk it temporarily engineering altered take involved
marketing increasingly n't protest the tower trading stands district further bid
six members engineers ms. above market of junk results offering about a transformed
concerns i think third-quarter she split the specific favored a truce in california
a federal investment in notes for tokyo requirement the coming scheduled earnings
ltd. 's basic to an me indicated when the burt 's \texttt{<unk>} associates changes in average
\end{quote}

\subsection*{A.4 WikiText-2 (Word-Level Transformer)}
\begin{quote}
The history of machine learning begins the thirty month Luis film Fingal ``Slayer Kingdom''
at the ’ well produced periodic being commented violence script disagree the microlight
third southern energy comparing was believe helicopter report largely everybody occurred
a goddesses bottom Psittacosaurus it realm 1918 Slayer team were Associated Great At a
Philips which Music more comes Star same guidance gods University and often @-@ their dies
haunting at the crack 1936 / to include discrimination mission deities. The Greek States
Reconnaissance was named up affected also needing splendid 203 in phenomena, and
performance of note had Station was strain 1 at his
\end{quote}

\subsection*{A.5 Tiny Shakespeare --- Transformer (Baseline Best Model)}
\begin{quote}
HAMLET: \\
In stry Iawath. \\[4pt]
KING LENM: \\
Not wands-- \\[4pt]
GAEULE: \\
No IZHARD IUCEZIO: \\
Come. \\[4pt]
LUCERj-tpestad fare, dea
\end{quote}

\subsection*{A.6 Tiny Shakespeare --- Self-Attention (Single Layer)}
\begin{quote}
HAMLET: \\
KI: \\
I'si a I for ty \\
That is loomt malot ithat tetral, fat reo. \\[4pt]
BONNGTERIV: \\
Ye, love blicestuee los
\end{quote}

\subsection*{A.7 Tiny Shakespeare --- MLP}
\begin{quote}
HAMLET: \\
Cer reen nat om the bo, the nreiv resse, \\
Thy thur a \\
yhun toret ofenty nowel iut fur: \\
Fh gicit Gle.
\end{quote}

\subsection*{A.8 Tiny Shakespeare --- Linear Model}
\begin{quote}
HAMLET: \\
ore is ,is t ae in the t er the d ine she , io d dit ur wiee, in  ie sine dound eall with drou de
\end{quote}

\end{document}